\documentclass[10pt]{article}
\usepackage[letterpaper]{geometry}
\usepackage{hicss}
\usepackage{times}
\usepackage[none]{hyphenat}
\usepackage{url}
\usepackage{latexsym}
\usepackage{indentfirst}
\usepackage{graphicx}
\graphicspath{{figures/}}
\usepackage[
    style=apa, citestyle=numeric
  ]{biblatex}
\usepackage[T1]{fontenc}

\usepackage{graphicx}
\usepackage{algorithm}
\usepackage{algorithmic}
\usepackage{booktabs}
\usepackage{amsmath}
\usepackage{xcolor, soul}
\colorlet{LightOrange}{orange!30}
\sethlcolor{LightOrange}
\usepackage{multirow}
\usepackage{amssymb}

\DeclareMathOperator*{\argmax}{arg\,max}
\DeclareMathOperator*{\argmin}{arg\,min}

\usepackage{graphicx}
\graphicspath{ {./figures/}}

\def\our{HebbCL}

\title{Hebbian Continual Representation Learning}

\author{Paweł Morawiecki \\
  Polish Academy of Sciences \\ Institute of Computer Science \\
  {\underline{ pawel.morawiecki@gmail.com}} \\ \\
  Andrii Krutsylo \\
  Polish Academy of Sciences \\ Institute of Computer Science \\
  {\underline{ akrutsyl@office.ibspan.waw.pl} } \\ \And
  Maciej Wołczyk \\
  Faculty of Mathematics and Computer Science \\ Jagiellonian University \\
  {\underline{ maciej.wolczyk@gmail.com} } \\ \\
  Marek Śmieja \\
  Faculty of Mathematics and Computer Science \\ Jagiellonian University \\
  {\underline{ smieja.marek@gmail.com} } \\ }

\date{}

\begin{document}
\maketitle
\begin{abstract}
Continual Learning aims to bring machine learning into a more realistic scenario, where tasks are learned sequentially and the i.i.d. assumption is not preserved. Although this setting is natural for biological systems, it proves very difficult for machine learning models such as artificial neural networks. To reduce this performance gap, we investigate the question whether biologically inspired Hebbian learning is useful for tackling continual challenges. In particular, we highlight a realistic and often overlooked unsupervised setting, where the learner has to build representations without any supervision. By combining sparse neural networks with Hebbian learning principle, we build a simple yet effective alternative (\our{}) to typical neural network models trained via the gradient descent. Due to Hebbian learning, the network have easily interpretable weights, which might be essential in critical application such as security or healthcare. We demonstrate the efficacy of \our{} in an unsupervised learning setting applied to MNIST and Omniglot datasets. We also adapt the algorithm to the supervised scenario and obtain promising results in the class-incremental learning.
\end{abstract}

\subsubsection*{Keywords:}

Continual Learning, interpretable neural network, Hebbian learning, unsupervised learning

\section{Introduction}

\begin{figure}[!t]
\includegraphics[width=0.5\textwidth]{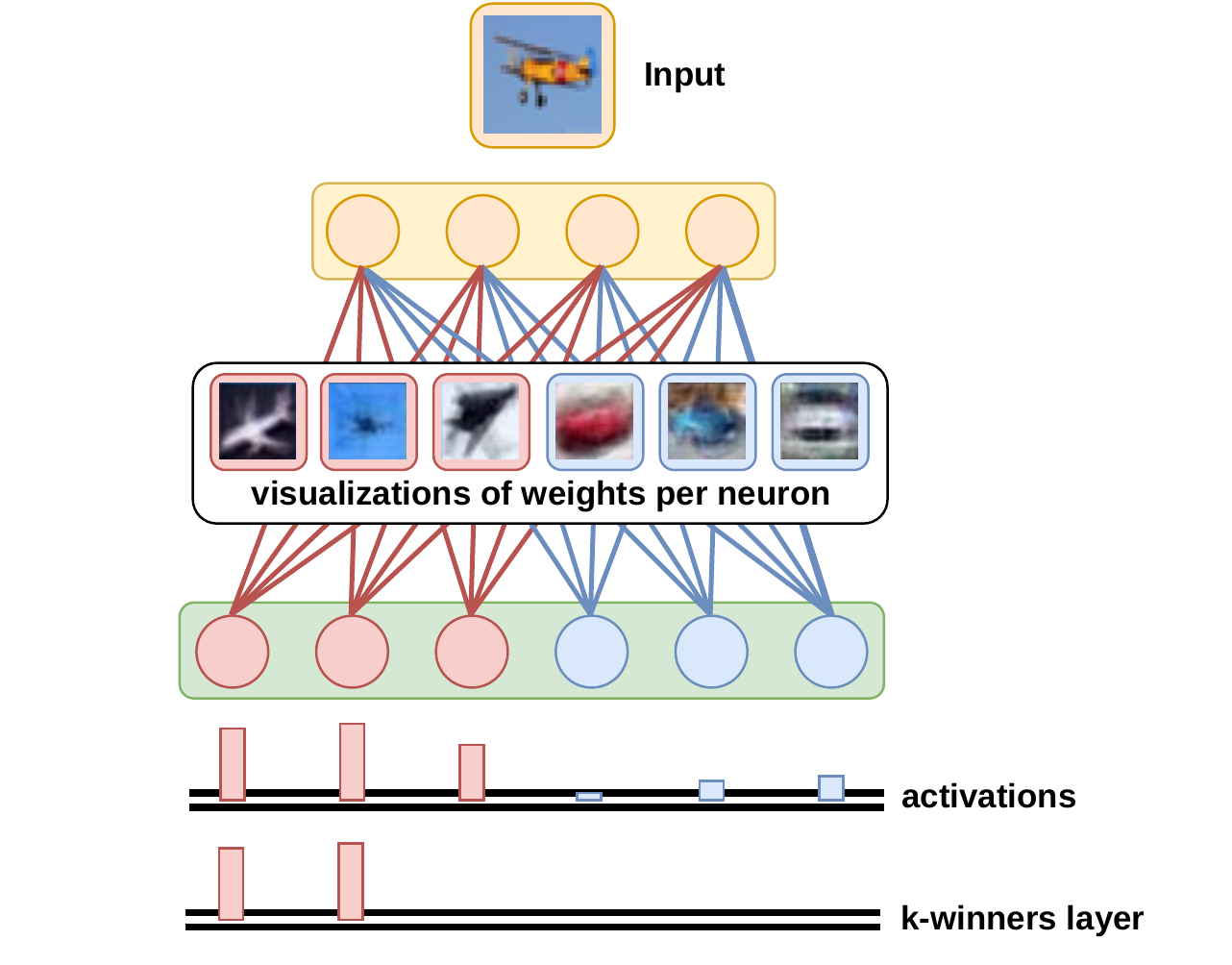}
\caption{A one-layer network trained by the unsupervised \our{} algorithm. Each visualization is a 3x32x32 weights of a neuron fully connected to the input layer. The more the neuron's weights are similar to the input sample, the more it will be activated. The final output is obtained by applying $k$-winners layer to the activation vector. Visualization of weights show that red neurons should recognize airplanes, whereas blue ones activate with a car as an input.}
\label{fig:introduction}
\end{figure} 

Humans are able to learn new tasks in a dynamic sequential manner and continuously adapt to new domains. Previously learned knowledge and skills are not only preserved and not forgotten, but the gained experiences are used to build representations for learning future tasks. Although artificial neural networks have been inspired by the biological neural systems, typical learning algorithms fail to remember and reuse learnt knowledge without introducing additional mechanisms \cite{kirkpatrick2017overcoming}. To mitigate catastrophic forgetting, neural networks repeatedly process a fixed dataset several times in a shuffled order. If we aim to construct general artificial intelligence, which solves lifelong problems, then such a scenario is highly unrealistic.

Continual Learning (CL) focuses on training neural network models from non-stationary and non-i.i.d. samples, where tasks/classes appear in a sequential manner \cite{van2019three}. Several deep learning approaches to CL have been recently introduced with a particular focus on supervised setting \cite{van2019three,hsu2018re,lomonaco2017core50}. Although in most cases biological systems are trained without supervision, the unsupervised CL has been mostly overlooked and only a few methods have been examined \cite{CURL,lee2020neural}.

In this paper, we address the unsupervised CL scenario, in which labels and boundaries between tasks are not provided. Our model builds a representation in a continual setting, which carries the information about present and past tasks. Obtained representation can be later used for downstream tasks such as clustering or classification. 

We develop our algorithm (HebbCL) relying on a few, biologically-inspired principles: sparsity of connections and activations, local Hebbian learning and a dynamic expansion of a network (adding new neurons during the training). Additionally, to limit the catastrophic forgetting we isolate (freeze) the weights, which carry meaningful information. Even though we focus on the unsupervised CL being the most challenging setting, our model can also be applied to the supervised case.    

The proposed learning algorithm leads to the interpretable weights of the network, easy to visualize and inspect. Such a feature might be essential at some critical applications including human safety and healthcare.

We conduct extensive experiments to evaluate the properties of the obtained representation in both unsupervised and supervised cases. For the unsupervised setting, we apply \our{} to MNIST and Omniglot datasets and evaluate the representation with cluster accuracy and k-Nearest Neighbour error. For most settings, we manage to improve state-of-the-art for these datasets. In the supervised setting, we work in the class-incremental learning setting and show promising results even for a harder dataset such as CIFAR-10. Finally, we carry out ablation study demonstrating the importance of key ingredients of \our{}.  

\section{Related Work}

This work lies at the intersection of Hebbian learning and continual learning and as such we describe related works in both of these areas.

\subsection{Hebbian Learning}

 Although backpropagation has proven to be an efficient method for calculating gradients in artificial neural networks, it is widely believed that it cannot be implemented in a brain \cite{bengio2015towards}. Because of that, several more biologically plausible alternatives were proposed, many of them inspired by the Hebbian rule of local learning which states that changes in synaptic strength should be only dependent on the presynaptic and postsynaptic neurons \cite{hebb2005organization}. 
 
 The most closely related work and an inspiration for our algorithm is the Hierarchical Temporal Memory (HTM) framework, where both sparsity and Hebbian learning are essential \cite{HTM}. In our paper we apply a simplified version of the Krotov-Hopfield learning rule for unsupervised training \cite{krotov2019unsupervised}. This rule was previously used with success for the image retrieval problem \cite{ryali2020bio} and embedding learning \cite{liang2021can}. 
 
 A separate line of research incorporates Hebbian principle to a typical, gradient-based training \cite{Halvagal2022.03.17.484712,Thangarasa2020EnablingCL}.

\subsection{Continual Learning} 

Continual learning (CL) studies how to learn from non-stationary streams of data, where the i.i.d. assumption is not satisfied. This setting approximates learning conditions in the real world, where the environment is constantly changing~\cite{hadsell2020embracing}. Although biological systems perform very well in such scenarios, artificial neural networks which excel in tasks with i.i.d. data have been shown to generally fail in the face of non-stationarity. In particular, the catastrophic forgetting problem appears, i.e. the performance on previous tasks rapidly deteriorates. This is an important difference between human-learning and artificial learning which has to be addressed in case we want to approach human-level training.

In the recent years the field of CL has grown considerably and numerous methods for overcoming catastrophic forgetting were proposed. These approaches can be divided into three major families of methods: regularization-based, replay-based and parameter isolation-based \cite{delange2021continual}, where each is inspired by different mechanisms of biological brains. Regularization-based methods \cite{li2017learning,zenke2017continual,aljundi2018memory} aim to penalize changes in the network which would reduce the performance on the previously seen tasks. For example, Elastic Weight Consolidation (EWC) \cite{kirkpatrick2017overcoming}, one of the standard methods CL approximates the second-order curvature of the loss function of the previous task and uses it as a penalization term when training on a new task. This approach can be linked to the synaptic mechanisms of the biological brains \cite{wixted2004psychology}. Parameter isolation methods \cite{mallya2018packnet,rusu2016progressive,schwarz2018progress} take this approach further by completely freezing parts of the network and introducing modularity to the structure of the model. For example, PackNet \cite{mallya2018packnet} freezes a small fraction of weights crucial for each task, and Progressive Networks \cite{rusu2016progressive} introduce new neurons for each task. Similarly, modularity seems to be an important part of biological neural systems \cite{meunier2009hierarchical}. Finally, replay-based methods \cite{chaudhry2019tiny,chaudhry2018efficient,buzzega2020dark,aljundi2018memory} save examples from the past for further retraining which allows them to maintain performance on previous tasks, relating to the concept of human memory.

As such, the field of CL is heavily influenced by biological solutions. However, most of the existing literature focuses on the task of supervised CL \cite{van2019three,hsu2018re,lomonaco2017core50}, and particularly the setting of image classification. Although there have been some approaches to address other domains and settings \cite{wolczyk2021continual,biesialska2020continual,nekoei2021continuous}, the research on unsupervised learning is relatively scarce. We argue that this setting is especially important, as humans rarely learn from clearly defined tasks with labels. In practice, biological learning is less structured and lacks supervision. This is why we focus on unsupervised CL. Although the related work here is more scarce, promising approaches include CURL \cite{CURL} and CN-DPM \cite{lee2020neural}, which focus on building mixtures of experts or latent distributions which specialize in parts of the data. In this work, we show that a more biologically plausible approach achieves competitive results in practice.

\section{\our{} Algorithm}

\begin{algorithm}[!t]
\caption{Unsupervised HebbCL algorithm} \label{Alg:train}
\begin{algorithmic}
\STATE Input: network (weights $W$), minibatch of training examples $X$
\STATE Hyperparameters: learning rate $\epsilon$, threshold $t$
\STATE Output: updated weights $W$
\STATE
\FOR {all $x_i \in X$}
    \STATE $m \gets \argmax (Wx_i)$ \hfill\COMMENT{identify the highest activation}
    \STATE $\Delta w \gets x_i - W_m$ \hfill\COMMENT{difference between input and weight vector}
    \STATE $W_m \gets W_m +\epsilon\Delta w$ \hfill\COMMENT{update weights}
\ENDFOR
\STATE $W_m \gets W_m/ \phi$
\hfill\COMMENT{normalize weights}
\STATE
\FOR {all row vectors $W_j \in W$}
    \FOR {all $x_i \in X$}
        \STATE $distances[i] \gets  d^2(W_j,x_i) \: / \: ||x_i||_1$ \hfill\COMMENT{calculate normalized Euclidean distance}
    \ENDFOR    
    \STATE $m \gets \argmin(distances)$ \hfill\COMMENT{identify the shortest distance}
    \IF{$distance[m] < t$}  
        \STATE {freeze $W_j$} \hfill\COMMENT{freeze a row vector of weights}
        \STATE {add a new neuron} \hfill\COMMENT{add a new row vector of weights}
    \ENDIF    
 
\ENDFOR
\STATE
\STATE \textbf{return} $W$

\end{algorithmic}
\end{algorithm}

\begin{algorithm}[!t]
\caption{Supervised HebbCL algorithm} \label{Alg:supervised}
\begin{algorithmic}
\STATE Input: network (weights $W$), training examples $X$ from a single class $C_j$
\STATE Hyperparameters: learning rate $\epsilon$, EPOCHS, a number of new neurons $n$
\STATE Output: updated weights $W$
\STATE
\STATE {denote unfrozen neurons as neurons belonging to class $C_j$}
\STATE
\FOR {e in range(EPOCHS)}
\FOR {all minibatches $B_k$}
\FOR {all $x_i \in B_k$}
    \STATE $m \gets \argmax (Wx_i)$ \hfill\COMMENT{identify the highest activation}
    \STATE $\Delta w \gets x_i - W_m$ \hfill\COMMENT{difference between input and weight vector}
    \STATE $W_m \gets W_m +\epsilon\Delta w$ \hfill\COMMENT{update weights}
\ENDFOR
\STATE $W_m \gets W_m / \phi$
\hfill\COMMENT{normalize weights}
\ENDFOR
\ENDFOR
\STATE

\STATE {freeze $W$} \hfill\COMMENT{freeze all the weights}
\STATE {expand the network with $n$ new neurons} \hfill\COMMENT{add $n$ new row vectors to $W$}
\STATE \textbf{return} $W$

\end{algorithmic}
\end{algorithm}


Throughout the paper we work in the class-incremental setting, that is classes are exposed to the model sequentially and during the inference time a task ID is not available. For the unsupervised setting --- our main point of interest --- we do not pass any knowledge of task boundaries to the learning algorithm. This makes the problem much more challenging than the standard benchmarks used for CL \cite{van2019three}, which usually provide labels, task ID and the task boundaries. 

\subsection{Algorithm overview}

We develop the learning algorithm following a few biologically inspired principles. The first one is the local Hebbian rule for updating the weights. In a typical gradient-based training all the weights are updated (to a various degree), even those which are essential for previously learned tasks. In the long term this causes the catastrophic forgetting. Having local Hebbian weight changes plus freezing meaningful weights should limit the catastrophic forgetting. We also introduce a dynamic expansion of the network (adding new neurons) to balance frozen neurons and maintain the learning capacity of the network. Finally, we make the representation sparse, which turns out to be useful for downstream tasks such as clustering and classification.  

\our{} builds a representation using a single, wide, fully-connected layer. Such the architecture has been proven to be successful for Hebbian learning in other contexts \cite{krotov2019unsupervised,liang2021can}. Hebbian learning can also be applied to deep networks but the accuracy of the network does not improve or even drops \cite{Hebbian_deep}. Any improvements on this research problem would also benefit \our{}.

From the high-level perspective, \our{} training consists of two main steps executed iteratively:
\begin{enumerate}
    \item Update the weights $W$ with the simplified Krotov-Hopfield rule
    \item If any neuron converged:
    \begin{enumerate}
    \item Freeze weights of converged neuron 
    \item Add a new neuron (expand $W$) 
    \end{enumerate}
\end{enumerate}

Once the training is over, the representation vector $y$ for the corresponding input vector $x$ is constructed by applying
\begin{equation}
\label{eq1}
y = f(Wx),
\end{equation}
where $W$ is a matrix representing network weights and $f$ is a function, which zeros all but $k$ highest activations (also known as `k-winners takes all' function)

\subsection{Unsupervised HebbCL}

We train the network in a CL fashion, without any knowledge of boundaries between classes of training examples. Training examples are processed in minibatches and these steps are executed for each minibatch. Steps 2(a) and 2(b) are conditional, that is weights are being frozen and a neuron is added only when certain conditions are matched. We give more details on this later in the section. 

The weights are updated by a local Hebbian rule. We do not use any labels or a global cost function typical for gradient-based optimization such as the backpropagation algorithm. Concretely, we apply the (slightly simplified) Krotov-Hopfield rule \cite{krotov2019unsupervised}. First, for a given training example $x$, we calculate the activation vector $a=Wx$ and identify the index $m$ of the most active neuron $a_m$, i.e.:
\begin{equation}
m  = \argmax_{m'} \, (Wx)_{m'}.
\end{equation}

Let us denote the weights connected to the most active neuron by $W_m$ ($m$-th row in the matrix $W$). Then the weights update from time step $i$ to $i+1$ is defined as follows:
\begin{equation}
\Delta w = x - W^{i}_m,
\end{equation}
\begin{equation}
W^{i+1}_m = W^{i}_m +\epsilon\Delta w/ \phi
\end{equation}
where $\epsilon$ is a learning rate and $\phi$ is a normalization factor equals to the largest absolute value of a weight in $W$. 

What drives the update rule is the difference between the input and the weights, $\Delta w = x - W_m$. As the training proceeds, the difference $\Delta w$ becomes smaller and weights $W_i$ converge towards a particular input pattern.      
 
To limit the catastrophic forgetting, we conditionally freeze some weights. Intuitively, we want to freeze those rows in $W$, for which the convergence to an input pattern is achieved and $\Delta w = x - W_j$ is very small. We calculate the squared Euclidean distance between $x$ and each $W_j$ and normalize it by the $l_1$ norm of $x$.

\begin{equation}
distance =  d^2(W_j,x) \: / \: ||x||_1
\end{equation}

 If $distance$ is smaller than a threshold $t$ (hyperparameter), then we freeze weights $W_j$ and add a new neuron to the representation $y$. (Threshold $t$ is determined experimentally on a validation set.) 
 
 New neurons are added to maintain the learning capacity of the network, otherwise there would be no trainable weights at some point of the training. Algorithm 1 shows the complete learning algorithm for a given minibatch.

Loops in the algorithm can be efficiently parallelized and implemented as a GPU-friendly code.

\subsection{Supervised HebbCL}

The proposed algorithm can be easily adapted to the supervised setting. If labels are known during training, they can be used to simplify freezing and expansion steps. 

Let us recall that in the class-incremental setting, the classes/tasks appear sequentially. Once the learning of a given class $C_j$ is over, we freeze all the weights and add a new portion of neurons for an upcoming class. This is in contrast to the unsupervised variant, where we do not have information about classes in training. To maintain the learning capability, we  conditionally freeze and expand only one neuron at a time. Algorithm 2 shows the pseudocode for the supervised \our{}.

During the training subsequent group of neurons are assigned to a given class. At the inference time, we sum activations for each group of neurons and the predicted class is the one with the highest total value.

\section{Experiments}
\subsection{Continual Unsupervised Representation Learning}

\begin{figure}[!t]
\includegraphics[width=0.5\textwidth]{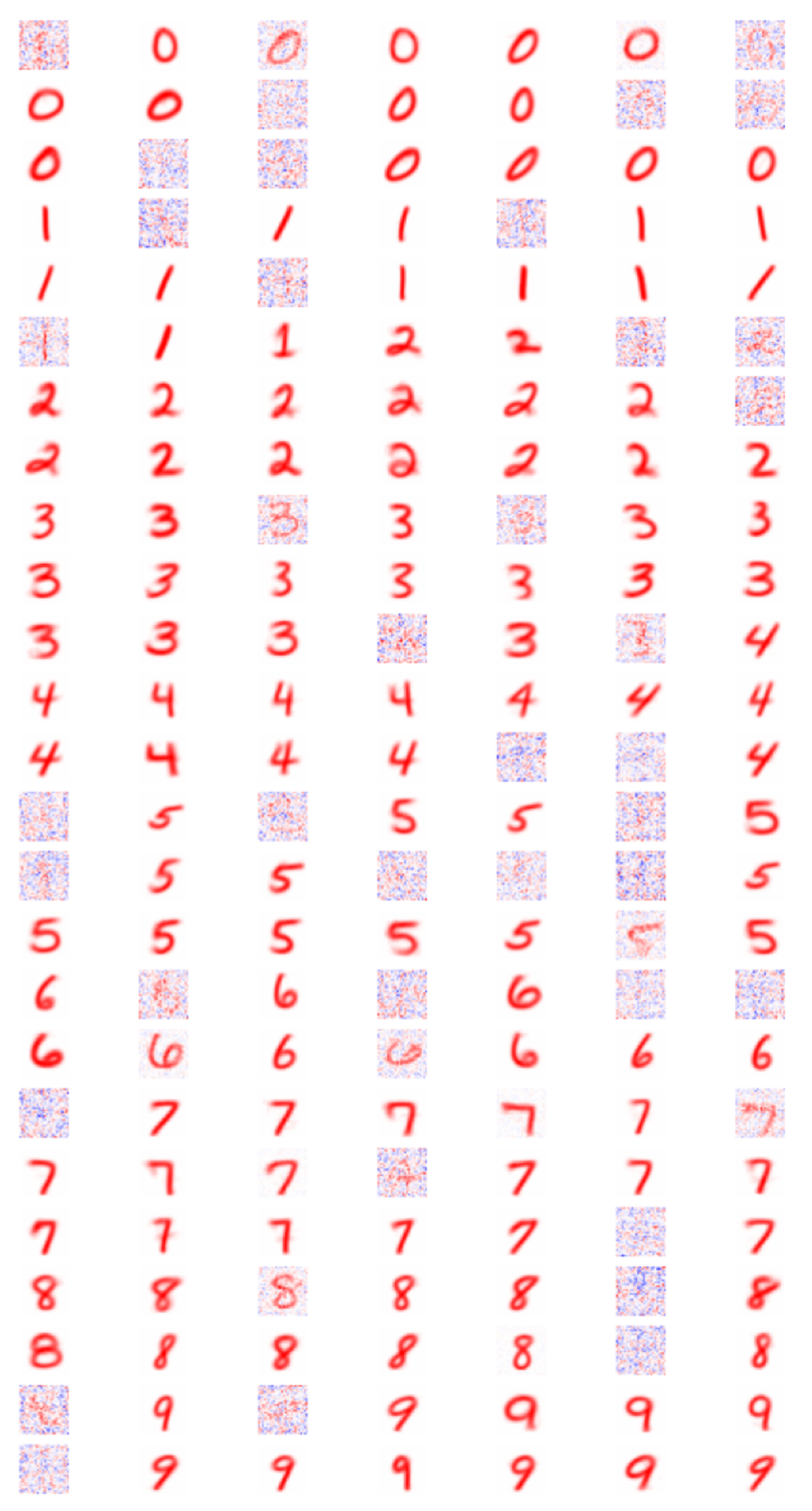}
\caption{Visualization of weights trained on MNIST. Each square represents 28x28 weights related to a given neuron.}
\label{fig:vis_MNIST_unsupervised}
\end{figure}

\begin{table*}[!t]
\centering
\caption{The average cluster accuracy and 10-Nearest Neighbours (10-NN) error across all five tasks of the split MNIST and Omiglot protocols, evaluated after learning the whole sequence. Each value is the average of five runs (with standard deviations).}
\label{tab:unsupervised}
\scalebox{0.88}{
\begin{tabular}{c|ccc|ccc} 
\toprule
\bf{Benchmark} & \multicolumn{3}{c|}{\bf{MNIST}} & \multicolumn{3}{c}{\bf{Omniglot}} \\
Method & \# clusters & Cluster acc (\%)$\uparrow$ & $10$-NN error (\%)$\downarrow$ & \# clusters & Cluster acc (\%)$\uparrow$ & $10$-NN error (\%)$\downarrow$  \\
\midrule

\our{} (our) & $25$         & $74.23_{\pm 0.67}$  & \hl{$6.48_{\pm 0.20}$}                & $50$  & $14.07_{\pm 0.34}$  & \hl{$71.62_{\pm 0.46}$} \\ 
          & $50$         & \hl{$78.35_{\pm 0.73}$} & \hl{$6.48_{\pm 0.20}$}                & $100$ & \hl{$16.46_{\pm 0.25}$}  &  \hl{$71.62_{\pm 0.46}$}\\ 

\midrule
CURL w/ MGR & $25.20_{\pm 2.23}$ & $77.74_{\pm 1.37}$ & $6.29_{\pm 0.50}$ & $101.20_{\pm 8.45}$ & $13.21_{\pm 0.53}$ & $76.34_{\pm 1.10}$ \\
CURL w/o MGR & $55.80_{\pm 1.94}$ & $45.35_{\pm 1.50}$ & $17.46_{\pm 1.25}$ & $189.60_{\pm 9.75}$ & $13.36_{\pm 1.06}$ & $81.91_{\pm 1.36}$ \\

\bottomrule
\end{tabular}}
\end{table*}

\begin{figure}[!t]
\includegraphics[width=0.491\textwidth]{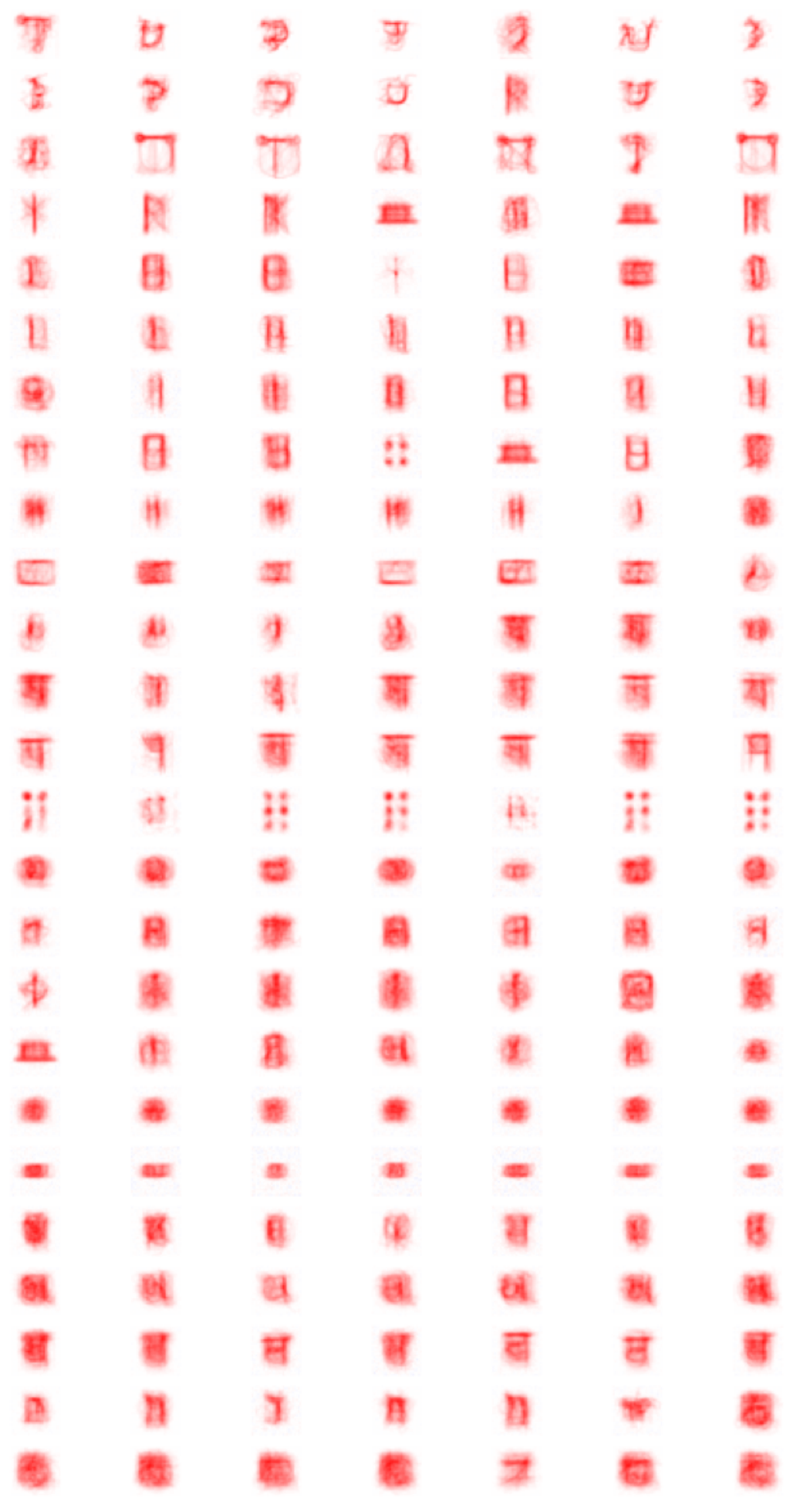}
\caption{Visualization of weights trained on Omniglot. Each square represents 105x105 weights related to a given neuron.}
\label{fig:vis_Omniglot_unsupervised}
\end{figure} 

\begin{table}[!t]
\centering
\caption{The average accuracy across all five tasks of the split MNIST and split CIFAR-10 protocols, evaluated after learning the whole sequence of tasks. Each value is the average of five runs (with standard deviations).}
\label{tab:results_mnist_cifar}
\setlength{\tabcolsep}{2pt}
\begin{tabular}{lllcc} 

\toprule
Method                &  & MNIST & CIFAR-10       \\
\midrule
SGD                   &  & $19.01 \pm 0.04$ & $15.56 \pm 5.08$  \\
L2                    &  & $18.88 \pm 0.18$  & $16.31 \pm 3.52$ \\ 
\midrule
EWC \cite{kirkpatrick2017overcoming}                   &  & $18.90 \pm 0.06$ & $11.83 \pm 4.10$  \\
                                            Online EWC \cite{schwarz2018progress}            &  & $18.89 \pm 0.07$ & $15.49 \pm 5.20$  \\
                                            Synaptic Intelligence \cite{zenke2017continual} &  & $17.94 \pm 0.57$ & $17.38 \pm 4.13$   \\
                                            MAS \cite{aljundi2018memory}                   &  & $17.38 \pm 4.19$  & $11.79 \pm 4.01$  \\
                                            LwF \cite{li2017learning}                   &  & $49.37 \pm 0.68$  & $13.89 \pm 5.33$  \\
                                                            EfficientPackNet \cite{schwarz2021powerpropagation}         &  & \hl{$99.42 \pm 0.15$} & ---   \\ \midrule            \our{} (our)        &  & $93.25 \pm 0.38$  & \hl{$40.91 \pm 0.30$}  \\

\bottomrule
\end{tabular}
\end{table}

\begin{figure}[!t]
\includegraphics[width=0.47\textwidth]{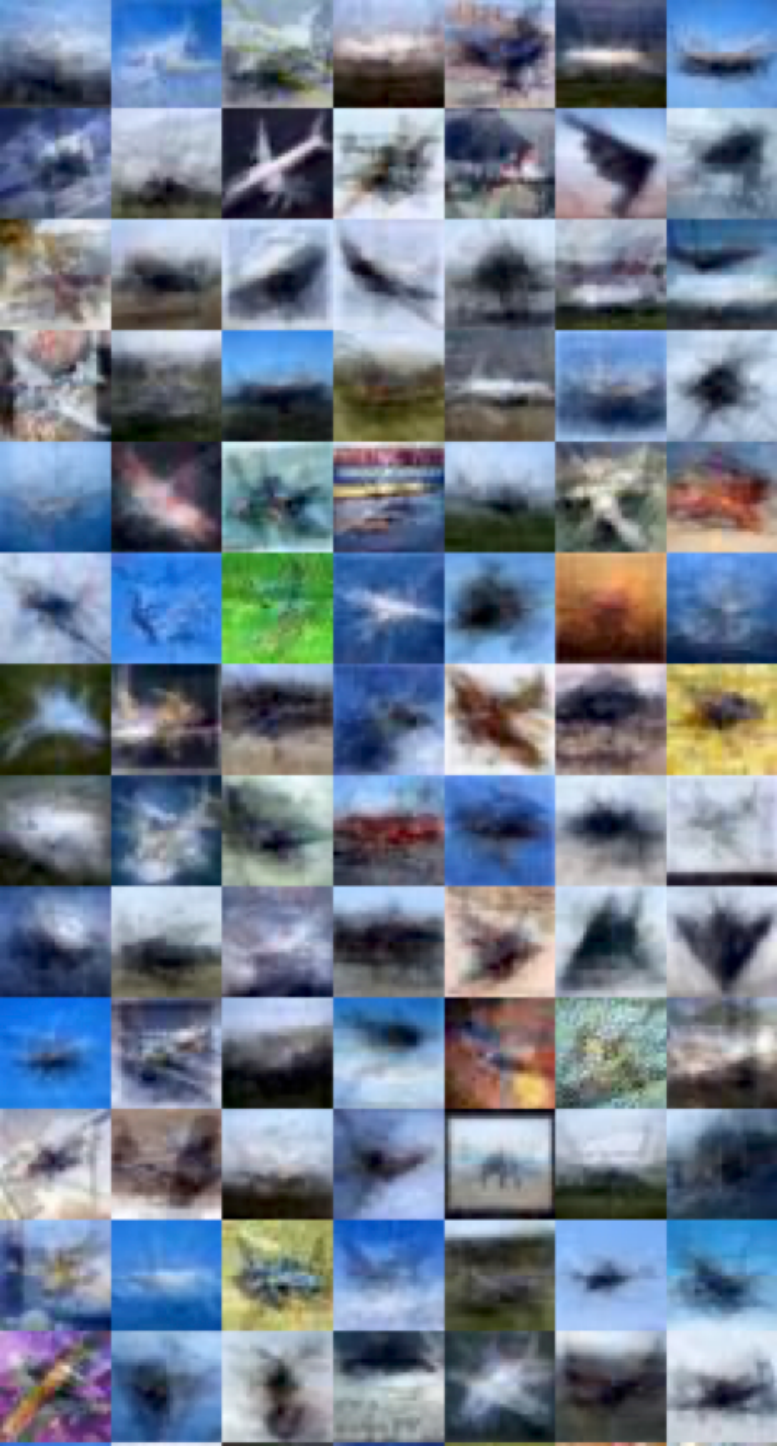}
\caption{Visualization of weights trained on all tasks of CIFAR-10 in the supervised setting. Each square represents 32x32x3 weights connected to a given neuron. Here we show only neurons related to the first class (airplanes) with possible influence of other classes.}
\label{fig:vis_cifar_supervised}
\end{figure} 

\begin{figure*}[!t]
\centering
\includegraphics[width=1.0\textwidth]{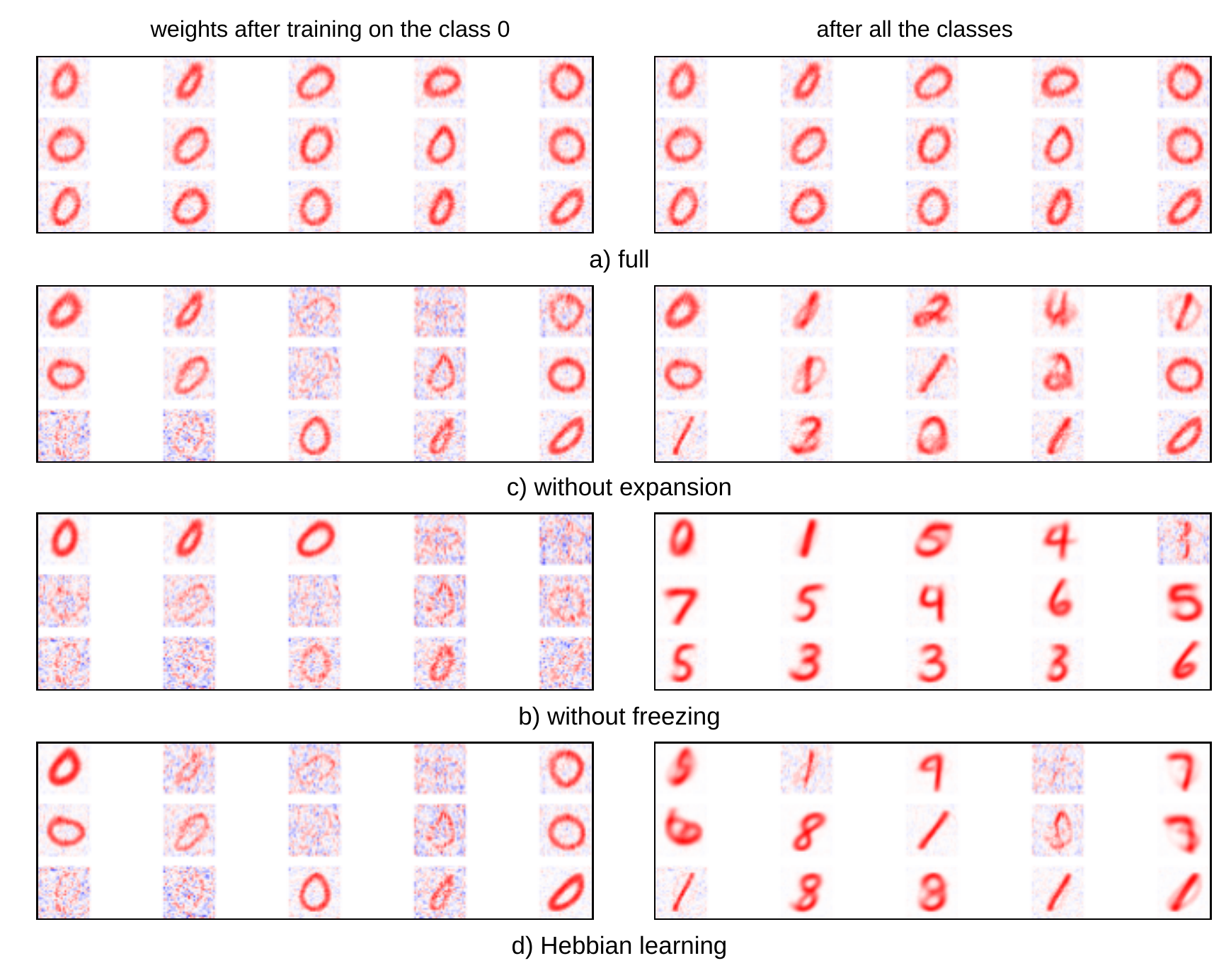}
\caption{Investigation of catastrophic forgetting for different variants of \our{}. Figures show visualizations of weights after training on the first class and the very same weights after a complete training on all classes. Clearly, in variants without freezing b) and d) the weights have been forgot (overwritten) by more recent classes.}
\label{fig:ablation}
\end{figure*}

To examine the proposed algorithm, we follow the evaluation protocol proposed in \cite{CURL}, where they introduced a method for the continual unsupervised representation learning called CURL. Following this paper, we test \our{} on two datasets: MNIST \cite{MNIST} and Omniglot \cite{Omniglot}. For MNIST we use the default split between training and test sets. For Omniglot we treat 50 alphabets as 50 classes, where for each character (letter) 15 examples are given to the training set and 5 to the test set.

We compare the representations constructed by \our{} and CURL. For CURL we report the best results for two variants, that is with and without a generative replay. We investigate whether the tested methods cluster the same classes together in the representation space. For that purpose, we use two metrics to evaluate the representation: cluster accuracy and k-Nearest Neighbours (k-NN) error. The cluster accuracy is calculated as follows. First, we cluster the obtained representations of the input dataset with the k-means algorithm. Then, we assign each cluster to its most represented class (true labels are available). With these newly assigned (predicted) labels and true labels we calculate the cluster accuracy for the test set. k-NN error is a percentage of wrongly classified examples from the test set by the k-NN classifier. 

Table \ref{tab:unsupervised} summarizes the results. As expected, Omniglot, with 50 classes and much higher sample variance, turns out to be a more challenging dataset than MNIST.
We obtain better results than CURL for Omniglot for both metrics. For MNIST, CURL and \our{} reach similar performance. We also stress that memory requirements for our algorithm is a few times lower than for CURL. \our{} uses smaller network, e.g., for MNIST a single hidden layer with 500 neurons.

Fig. \ref{fig:vis_MNIST_unsupervised} shows the weights visualization after the training on the MNIST dataset. Most of weights contain meaningful information. Every class is represented and variety of samples is also reflected. Although there are some `noisy' weights, they do not affect the final representations, as their activations are zeroed anyway due to the $k$-winners layer .    

Omniglot has much higher sample variance within a class, which makes the problem more challenging for unsupervised learning methods. Generally, \mbox{\our{}} captures all the alphabets (classes), however, not every letter/sign can be clearly recognized (see Fig. \ref{fig:vis_Omniglot_unsupervised}).

\subsection{Continual Supervised Learning}

We evaluate the supervised \our{} on the split MNIST and split CIFAR-10 benchmarks, where the data are split into five tasks, each classifying between two classes and the model is trained on each task sequentially. At the inference time, we do not provide a task id, that is we work in the class-incremental learning scenario \cite{class-incremental}, which is the most challenging setting for supervised CL. 

Since our method does not use a replay buffer or any generative processes, we restrict comparison to the regularization-based and parameter-isolation methods. For split MNIST \our{} substantially outperforms the regularization methods and is close to best parameter-isolation approach such as EfficientPackNet \cite{EfficientPackNet}.
CIFAR-10 is a more complex dataset and we use a wider network with 2000-unit feature vector (compared to 640 for MNIST). Trained weights are still meaningful, easy to assign with a particular class, see Fig. \ref{fig:vis_cifar_supervised}. Here we also obtain substantially better accuracy than regularization-based methods (Table \ref{tab:results_mnist_cifar}). The authors of EfficientPackNet did not provide results for harder benchmarks (such as CIFAR) in the class-incremental setting.

%

\subsection{Ablation study}

\begin{table*}[!t]
\centering
\caption{Ablation study for \our{} performed on unsupervised setting. The sign \checkmark \  indicates which steps were used in the experiment - (H)ebbian learning, (F)reezing weights, (E)xpansion and (K)-winners}
\label{tab:ablation}
\scalebox{0.9}{
\begin{tabular}{cccc|cc|cc} 
\toprule
\multicolumn{4}{c}{\bf{Steps}} & \multicolumn{2}{|c|}{\bf{MNIST}} & \multicolumn{2}{c}{\bf{Omniglot}} \\
H & F & E & K & Cluster acc (\%)$\uparrow$ & $10$-NN error (\%)$\downarrow$ & Cluster acc (\%)$\uparrow$ & $10$-NN error (\%)$\downarrow$  \\
\midrule

\checkmark & \checkmark & \checkmark & \checkmark & $63.09_{\pm 1.53}$ & $6.48_{\pm 0.21}$ & $14.07_{\pm 0.34}$ & $71.62_{\pm 0.46}$ \\ 
\checkmark & \checkmark & & \checkmark & $28.96_{\pm 0.43}$ & $8.79_{\pm 0.14}$ & $9.30_{\pm 0.19}$  & $81.95_{\pm 0.31}$ \\
\checkmark & & & \checkmark & $48.56_{\pm 0.36}$ & $7.99_{\pm 0.20}$ & $14.50_{\pm 0.27}$ & $79.81_{\pm 0.49}$ \\
\checkmark & \checkmark & \checkmark & & $35.62_{\pm 0.14}$ & $6.14_{\pm 0.05}$ & $12.45_{\pm 0.21}$ & $71.92_{\pm 0.53}$ \\
\checkmark & & & & $32.2_{\pm 0.15}$ & $6.8_{\pm 0.03}$ & $14.42_{\pm 0.16}$ & $60.18_{\pm 0.22}$ \\

\bottomrule
\end{tabular}}
\end{table*}

\our{} have four core
elements --- Hebbian learning, freezing
weights, adding new neurons (expansion) and zeroing all but $k$ highest
activations. To assess the impact of these steps, we conduct an
ablation study for the unsupervised scenario measuring the cluster accuracy and the 10-NN error. A number of clusters for MNIST and Omniglot is 10 and 50, respectively. Table \ref{tab:ablation} indicates that removing one or
several steps from the algorithm reduces performance for both metrics. Figure \ref{fig:ablation} visualizes how changes are reflected in the generated representations. 

In a variant without expansion, all neurons are initially available for training, which causes the model to exhaust its capacity after first tasks.
The model cannot learn representations of new classes when most neurons
are frozen. In a variant without freezing, new classes tend to overwrite the stored representations, leading to forgetting earlier tasks. When both freezing and expansion are disabled, representations become less distinctive  
Interestingly, omitting the sparse representation (disabling $k$-winners layer) reduces the cluster accuracy greatly, particularly for MNIST.  

For Omniglot, the accuracy drop is less consistent. We hypothesise it is due to more complex nature of the dataset (50 classes, high sample variance) and the dynamics between elements of the algorithm is less predictable.  

\section{Conclusion}

We developed a method called \our{} to address the unsupervised CL problem, in which task labels and boundaries are unknown. The algorithm relies on a few biologically inspired principles: sparsity, Hebbian learning and dynamic growth of the network. For MNIST and Omniglot we improve state-of-the-art results in the unsupervised setting. We also adapted the method to the supervised scenario and showed competitive results. 

As our visualizations show, the proposed learning algorithm  leads to easily interpretable weights --- a desired feature in many critical applications.

We used shallow but wide networks (up to 6 mln parameters). However, for more complex datasets, a deeper network is necessary. The main challenge and an open problem is how to scale \our{} for deeper networks without losing desired properties for the CL. We leave this research question for future work.  

\addtolength{\textheight}{-.2cm} 


\printbibliography

\end{document}